\documentclass[letterpaper, 10 pt, conference]{ieeeconf} 

\usepackage{graphicx}
\usepackage{amsmath}
\usepackage{dsfont}
\usepackage{graphicx}
\usepackage{float}
\usepackage{pifont}
\usepackage{pgf}
\usepackage{tikz}
\usepackage{commath}
\usepackage{subfig}
\usepackage{epstopdf}
\usepackage[]{algorithm2e}
\usepackage{algorithmic}
\usepackage{ifthen}
\usepackage{subfig}
\usepackage{bbm}
\usepackage{dsfont}
\usepackage{amssymb}
\usepackage{txfonts}
\usepackage{mathbbol}
\usepackage{multirow}
\usepackage{lipsum}

\IEEEoverridecommandlockouts                              

\author{Rumit Kumar$^{1}$, Mahathi Bhargavapuri$^{2}$, Aditya M. Deshpande$^{1}$, \\ Siddharth Sridhar$^{1}$, 
Kelly Cohen$^{1}$, and Manish Kumar$^{1}$ 
\thanks{$^{1}$R. Kumar, A. M. Deshpande, S. Sridhar, K. Cohen and M. Kumar are with University of Cincinnati, Cincinnati, Ohio 45221, USA, {\tt\small Email: (kumarrt, deshpaad, sridhasd) @mail.uc.edu, (kelly.cohen, manish.kumar)@uc.edu }}
\thanks{$^{2}$M. Bhargavapuri is with Indian Institute of Technology, Kanpur, India {\tt\small Email: mahathi@iitk.ac.in}}}

\title{\Large \bf Quaternion Feedback Based Autonomous Control of a Quadcopter UAV \\ with Thrust Vectoring Rotors}

\newtheorem{lemma}{Lemma}

\begin{document}
\maketitle
\thispagestyle{empty}
\pagestyle{empty}

\begin{abstract}
In this paper, we present an autonomous flight controller for a quadcopter with thrust vectoring capabilities. This UAV falls in the category of multirotors with tilt-motion enabled rotors. Since the vehicle considered is over-actuated in nature, the dynamics and control allocation have to be analysed carefully. Moreover, the possibility of hovering at large attitude maneuvers of this novel vehicle requires singularity-free attitude control. Hence, quaternion state feedback is utilized to compute the control commands for the UAV motors while avoiding the gimbal lock condition experienced by Euler angle based controllers. The quaternion implementation also reduces the overall complexity of state estimation due to absence of trigonometric parameters. The quadcopter dynamic model and state space is utilized to design the attitude controller and control allocation for the UAV. The control allocation, in particular, is derived by linearizing the system about hover condition. This mathematical method renders the control allocation more accurate than existing approaches. Lyapunov stability analysis of the attitude controller is shown to prove global stability. The quaternion feedback attitude controller is commanded by an outer position controller loop which generates rotor-tilt and desired quaternions commands for the system. The performance of the UAV is evaluated by numerical simulations for tracking attitude step commands and for following a way-point navigation mission.
\end{abstract}
\section{Introduction}
The design advances in vertical takeoff and landing (VTOL) UAV research have led to the development of various types of quadcopter platforms. These design variants are based on the simple four propeller model to produce thrust for VTOL but utilize different methodologies for attitude control and navigation of the quadcopter in three-dimensional space. These methodologies can be implemented at flight software level as well as hardware level in the UAV. The hardware design advances include variable blade pitch quadcopters, tilt-rotor quadcopters, engine powered and re-configurable unmanned aerial systems. In a variable blade pitch quadcopter, motion and orientation control is achieved by the change in blade pitch angle of different rotors in various combinations. This UAV platform is capable of producing reverse thrust and inverted flights \cite{gupta2016flight} \cite{cutler2015analysis}. In a tilt-rotor UAV, the propeller motors are actuated to tilt about the arm connecting to the main quadcopter body using servo motors \cite{ryll2012modeling}. This UAV can follow tight trajectories and provide better disturbance rejection towards uncertainties during flight \cite{kumar2017tilting} \cite{kumar2017position}. This paper focuses on position and attitude control design for the tilt-rotor UAV using quaternion state feedback and accurate control allocation for the over-actuated system. Dynamic modeling and control design methods for the various flight modes of tilt-rotor quadcopter are discussed in  \cite{kumar2017tilting}, \cite{ali1}, \cite{ryll2015novel}, \cite{sridhar2017non} and \cite{sridhar2019tilt}. The work in \cite{oosedo2015flight} has shown that the tilt-rotor quadcopter can achieve large attitude angles. In recent works, Franchi \textit{et al.} in \cite{franchi2018full} have defined a general class of tilt-rotor UAV with laterally bounded input force for full-pose tracking. 
\begin{figure}[t]
	\centering
	\includegraphics[scale=0.30]{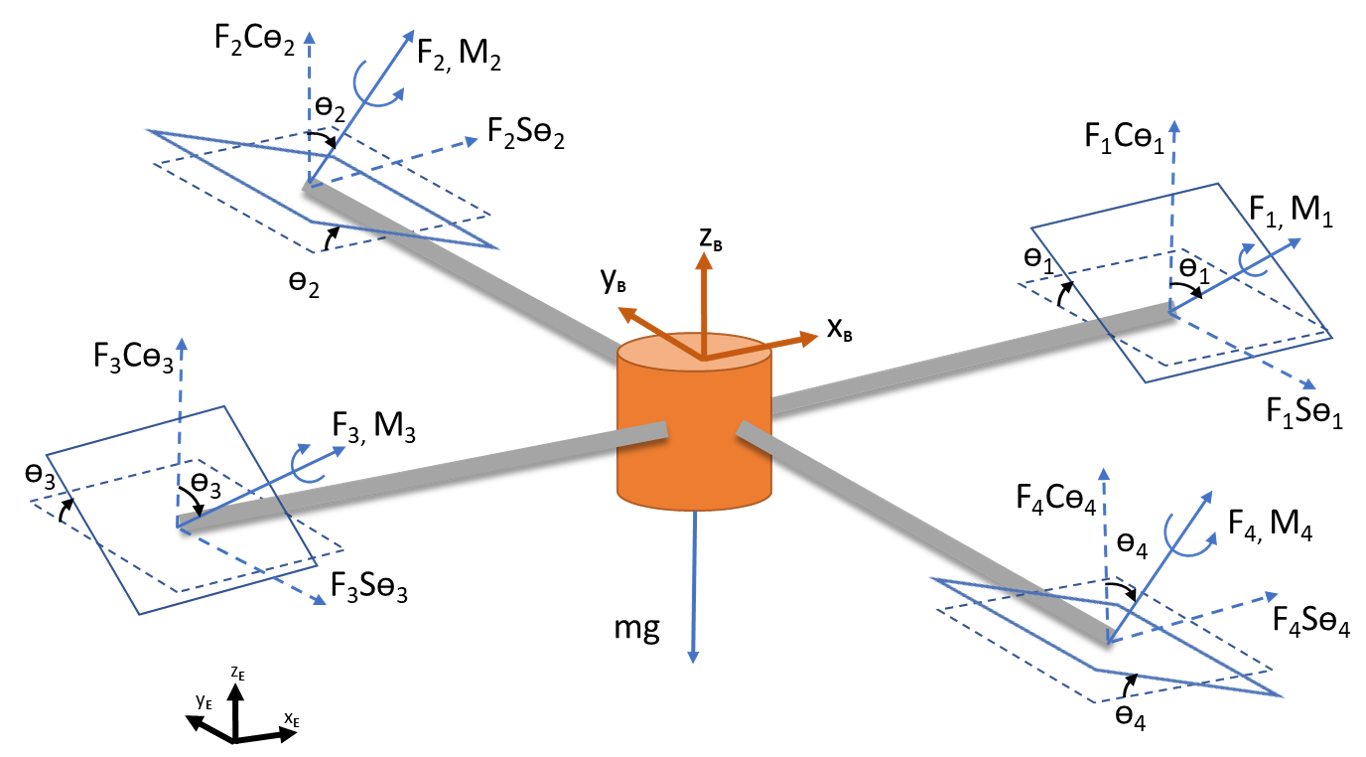}
    \caption{Tilt-rotor quadcopter free body diagram \cite{kumar2017tilting}}
    \label{fig1}
\end{figure}
Invernizzi \textit{et al.} in \cite{invernizzi2018trajectory} have utilized geometric control theory to develop control laws for tilt-rotor UAV. Similarly, fault-tolerant control in case of a single propeller or motor failure using tilt-rotor UAV are discussed in \cite{nemati2016stabilizing}\cite{kumar2018reconfigurable} \cite{sridhar2018fault}. These previous works on tilt-rotor quadcopter utilize Euler angles and direction cosine matrices formulation for developing flight controllers. From dynamics and control perspective, the attitude characteristics of any rigid body cannot be extracted by integrating the angular velocities in body frame because the Euler angles are defined in different frames and are only locally valid. So, the control engineers rely on integral solution of kinematic equations which are limited by inherent singularities and only a limiting solution for attitude can be obtained when two of the rotational axes coincide \cite{pamadi2015performance}. This phenomenon is termed as gimbal lock where, we can not distinguish between two degrees of freedom because those rotational axes coincide with each other \cite{fresk2013full}. Quaternion feedback based controller is a viable solution to overcome gimbal lock limitations as well as the complexity of estimating rotation matrices. The control applications of quaternion feedback based controller for multi-rotor UAV platforms are shown in \cite{cutler2015analysis}, \cite{stingu2009design}, \cite{carino2015quadrotor}, and \cite{fresk2015experimental}. Bhargavapuri \textit{et al.} in \cite{bhargavapuri2019vision} have used quaternion feedback based attitude controller for tilt-rotor quadcopter. The work  by Fresk \textit{et al.} in \cite{fresk2013full} and \cite{fresk2015experimental} for quaternion feedback based attitude controller for quadcopters is quite notable. Although, quaternion feedback controllers are developed in literature for traditional quadcopters, this work is one of the first attempts to develop quaternion feedback based attitude controller for an over-actuated tilt-rotor UAV. The detailed mathematical model for translational and rotational dynamics of the tilt-rotor UAV are presented. The moment equations are linearized using small perturbation theory or Taylor's series expansion to derive the necessary control allocation for the attitude control of the UAV. This is the first time such a mathematical approach is presented for derivation of control allocation for tilt-rotor UAVs. It has been shown that attitude control using quaternions is actually a regulation problem in terms of error quaternion. Further, Lyapunov stability analysis for quaternion control regulatory loop is shown to establish stability of the closed-loop system. The attitude controller is validated by numerical simulations and the inner quaternion attitude feedback loop is commanded by integrating an external position controller loop for achieving autonomous way-point navigation.

\section{Dynamic Model} \label{sec11}
This section provides a brief introduction to quaternions and the dynamic model including equations governing the translational and rotational motion of the UAV. A quaternion is a hypercomplex number in $\mathds{R}^4$, consisting of a scalar part and a vector part with three elements as shown in \eqref{eq4}.
\begin{eqnarray}
\mathbb{A} &=& \mathbb{A_0} + \mathbb{A_1}i + \mathbb{A_2}j + \mathbb{A_3}k \label{eq4}
\end{eqnarray}
where, $\mathbb{A_0}$ is the scalar part and $\mathbb{A_1, A_2, A_3}$ are the elements of the vector part ($\Bar{\mathbb{A}}$) of the quaternion.
Quaternion multiplication ($\mathbb{A}\bigotimes\mathbb{B}$), quaternion conjugate ($\mathbb{A}^{*}$), and quaternion normalize ($ \Hat{\mathbb{A}}$) are the main operations used in this work. The detailed explanation on quaternion operations can be found in \cite{stevens2015aircraft}. The expression of quaternion derivative for describing the vehicle attitude is shown in \eqref{eq8} and \eqref{eq9}.
\begin{eqnarray}
\dot{\mathbb{q}} &=& \frac{1}{2} \mathbb{q} \bigotimes \begin{bmatrix}
0 \\ \Omega
\end{bmatrix} \label{eq8} \\
\begin{bmatrix}
\dot{\mathbb{q_0}} \\ \dot{\mathbb{q_1}} \\ \dot{\mathbb{q_2}} \\ \dot{\mathbb{q_3}}
\end{bmatrix} &=& \frac{1}{2}
\begin{bmatrix}
0 & -p & -q & -r \\
p & 0 & r & -q \\
 q & -r & 0 & p\\
r & q & -p & 0
\end{bmatrix} 
\begin{bmatrix}
\mathbb{q_0} \\ \mathbb{q_1} \\ \mathbb{q_2} \\ \mathbb{q_3}
\end{bmatrix} \label{eq9} 
\end{eqnarray}
In equation \eqref{eq8}, $\Omega=\begin{bmatrix}p & q & r\end{bmatrix}^T$ represent the vehicle body rates and the quaternion can be estimated by integrating \eqref{eq8}. The normalize operation is used to convert a quaternion into unit quaternion as shown in \eqref{eq10}.
\begin{equation}
\Hat{\mathbb{q}} = \dfrac{\mathbb{q_0} + \mathbb{q_1}i + \mathbb{q_2}j + \mathbb{q_3}k}{\sqrt{\mathbb{q_0}^{2} + \mathbb{q_1}^{2} + \mathbb{q_2}^{2} + \mathbb{q_3}^{2}}} \label{eq10}
\end{equation}
The conversion from quaternion to Euler angles can be achieved by using the operation shown in \eqref{eq11}. 
\begin{equation}
\begin{bmatrix}
\phi \\ \theta \\ \psi
\end{bmatrix}
= \begin{bmatrix}
\mathrm{atan2}[2(\mathbb{\hat{q_0}}\mathbb{\hat{q_1}} + \mathbb{\hat{q_2}\hat{q_3}}), 1 - 2(\mathbb{\hat{q_1}}^2 + \mathbb{\hat{q_2}}^2)]\\
\mathrm{asin}[2(\mathbb{\hat{q_0}\hat{q_2} - \hat{q_3}\hat{q_1}})]\\
\mathrm{atan2}[2(\mathbb{\hat{q_0}\hat{q_3} + \hat{q_1}\hat{q_2}}), 1 - 2(\mathbb{\hat{q_2}}^2 + \mathbb{\hat{q_3}}^2)]
\end{bmatrix}
\label{eq11}
\end{equation}
The rotating propellers in the quadcopter produce forces and moments and the tilting motion of rotating propellers helps in force and moment vectoring as shown in Fig. \ref{fig1} \cite{kumar2017tilting}. The translational motion dynamics of the tilt-rotor quadcopter are described by equation \eqref{eq0}
\begin{eqnarray}
\begin{bmatrix} 
\ddot{x}\\\\
\ddot{y}\\\\ 
\ddot{z} 
\end{bmatrix}  
= \mathbb{q} \otimes
\begin{bmatrix} 
 \frac{F_2s\theta_2 + F_4s\theta_4}{m}\\\\
\frac{-F_1s\theta_1 - F_3s\theta_3}{m}\\\\
\frac{F_1c\theta_1 + F_2c\theta_2 + F_3c\theta_3 + F_4c\theta_4}{m}
\end{bmatrix}
\otimes
\mathbb{q^{*}}
-
\begin{bmatrix} 
 0 \\\\
 0\\\\
 g
\end{bmatrix} \label{eq0} 
\end{eqnarray}
where $\mathbb{q}$ is the quaternion that transform the acceleration vector from body frame to the inertial frame of reference. The $sine$ and $cosine$ angle terms are shown as $s\angle$ and $c\angle$ respectively. Further, $\theta_i$, $\forall{i}\in$ $\lbrace 1,2,3,4\rbrace$ are the rotor tilt angles, $m$ is the total mass of quadcopter, $g$ is the acceleration due to gravity, $\ddot{x}$, $\ddot{y}$ and $\ddot{z}$ are the linear accelerations in the world frame,  $F_i$, $\forall{i}\in$ $\lbrace 1,2,3,4\rbrace$ represent the propeller thrust forces. The rotational dynamics of the UAV are represented as effective torque along body axes as shown in \eqref{eq1} and \eqref{eq2}
\begin{eqnarray}
\dot{\Omega} &=& I^{-1} \medspace \medspace (\tau - \Omega \times I \medspace \Omega) \label{eq1}\\
\tau &=& \begin{bmatrix} 
l(F_2c\theta_2 - F_4c\theta_4)+ M_2s\theta_2 + M_4s\theta_4\\\\
l(F_3c\theta_3 - F_1c\theta_1)+ M_3s\theta_3 + M_1s\theta_1\\\\
l(-F_1s\theta_1 - F_2s\theta_2+ F_3s\theta_3 + F_4s\theta_4)\\
- M_1c\theta_1 + M_2c\theta_2 - M_3c\theta_3 + M_4c\theta_4
\end{bmatrix} \label{eq2}
\end{eqnarray}
where $I$  is the diagonal moment of inertia matrix with $[I_{xx}, \medspace I_{yy}, \medspace I_{zz}]$ as the diagonal elements. $M_i$, $\forall {i}\in \lbrace1,2,3,4\rbrace$ are the moments produced by the rotor drag.
The propeller thrust force $F_i$ and moment $M_i$ are related to the rotational angular speed $\omega_i$ of the $i$th rotor by \eqref{eq3}
\begin{equation}
F_i=k_f\omega_i^2, \qquad   M_i=k_m\omega_i^2, \quad \forall {i}\in \lbrace1,2,3,4\rbrace     \label{eq3}
\end{equation}
where $k_f$ and $k_m$ are force and moment coefficients, respectively \cite{grasp}. It should be noted that any variations in propeller speeds and rotor tilts lead to a resultant torque about body axes producing angular accelerations. The body rates can be obtained by integral solution of \eqref{eq1}. These rates can be estimated in an actual system using an inertial measurement unit (IMU) sensor. Further, the body rates can be utilized to estimate quaternions using \eqref{eq8}. 

\section{Controller Development}
\subsection{Attitude Control} \label{sec41}
In this section, the attitude controller of tilting rotor quadcopter is derived using quaternion feedback. The rotational dynamics of the system are linearized at a hovering state. At the hovering state, all propellers of the UAV are spinning at a nominal angular speed ($\omega_h$) and the rotors are tilted by ($\theta_h$) angle. The roll, pitch and yaw angles are all zero in the hovering state. This condition is related to a unit quaternion, thus the vehicle attitude can be represented by $[1 \thickspace \thickspace 0 \thickspace \thickspace 0 \thickspace \thickspace 0]^{T}$ quaternion. This assumption simplifies quaternion kinematics equation \eqref{eq9}. The quaternion rate and acceleration directly relate to body rates and accelerations of the system respectively and the control input torques will affect the quaternion vector elements as shown in \eqref{eq13}.
\begin{eqnarray}
\begin{bmatrix}
\dot{\mathbb{q_0}} \\ \dot{\mathbb{q_1}} \\ \dot{\mathbb{q_2}} \\ \dot{\mathbb{q_3}}
\end{bmatrix} &=& \frac{1}{2}
\begin{bmatrix}
0 \\ p \\ q \\ r
\end{bmatrix}; \qquad
\begin{bmatrix}
\ddot{\mathbb{q_0}} \\ \ddot{\mathbb{q_1}} \\ \ddot{\mathbb{q_2}} \\ \ddot{\mathbb{q_3}}
\end{bmatrix} = \frac{1}{2}
\begin{bmatrix}
0 \\ \dot{p} \\ \dot{q} \\ \dot{r} 
\end{bmatrix} \label{eq13}
\end{eqnarray}
We can solve \eqref{eq1} analytically using Taylor's series expansion. The rate multiplication Coriolis terms are very small and we ignore them to simplify the rotational dynamics \cite{pamadi2015performance}.
\begin{eqnarray}
I
\begin{bmatrix}
\dot{p} \\ \\ \dot{q} \\ \\ \dot{r}
\end{bmatrix}
&=& \begin{bmatrix} 
l(F_2c\theta_2 - F_4c\theta_4)+ M_2s\theta_2 + M_4s\theta_4\\\\
l(F_3c\theta_3 - F_1c\theta_1)+ M_3s\theta_3 + M_1s\theta_1\\\\
l(-F_1s\theta_1 - F_2s\theta_2+ F_3s\theta_3 + F_4s\theta_4)\\
- M_1c\theta_1 + M_2c\theta_2 - M_3c\theta_3 + M_4c\theta_4
\end{bmatrix} \label{eq14}
\end{eqnarray}
The solution for angular motion about $x_b$ axis is presented here and it can be generalized about $y_b$ and $z_b$ axes. Equation for angular motion about $x_b$ axis is shown in \eqref{eq15} using the force and moment values from equation \eqref{eq3}.
\begin{figure}[t]
	\centering
	\includegraphics[scale=0.5]{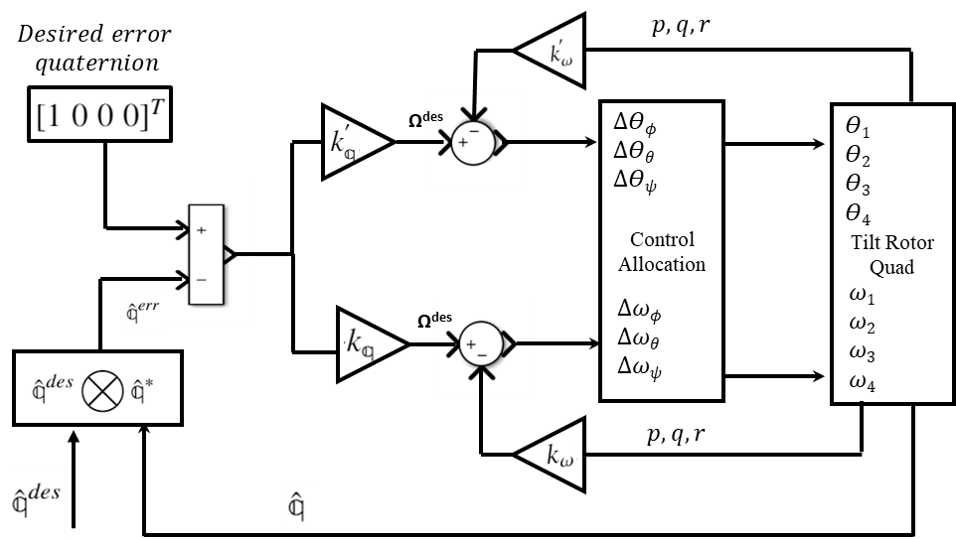}
    \caption{ Attitude control architecture} \vspace{-5mm}
    \label{fig2}
\end{figure} 
\begin{equation}
I_{xx} \medspace \dot{p} = lk_f\omega_2^{2}c\theta_2 - lk_f\omega_4^{2}c\theta_4+ k_m\omega_2^{2}s\theta_2 + k_m\omega_4^{2}s\theta_4 \label{eq15}
\end{equation}
Taylor's series expansion about the hover linearization point:
\begin{eqnarray}
\nonumber I_{xx} \medspace \delta\dot{p} = 2lk_f \omega_2 \delta \omega_2 c\theta_2 - lk_f\omega_2^{2}s\theta_2 \delta \theta_2 - 2lk_f\omega_4 \delta \omega_4 c\theta_4 \\ \nonumber + lk_f\omega_4^{2}s\theta_4 \delta \theta_4 + 2k_m\omega_2 \delta \omega_2 s\theta_2 + k_m\omega_2^{2} c\theta_2 \delta \theta_2 \\ \nonumber + 2k_m\omega_4 \delta \omega_4 s\theta_4 + k_m\omega_4^{2} c\theta_4 \delta \theta_4
\end{eqnarray}
\noindent Based on the assumption for linearization about hover condition $\omega_2 = \omega_4 = \omega_h$ and $\theta_2 = \theta_4 = \theta_h$.
\begin{eqnarray}
\nonumber I_{xx} \medspace \delta\dot{p} = 2lk_f \omega_h \delta \omega_2 c\theta_h - lk_f\omega_h^{2}s\theta_h \delta \theta_2 - 2lk_f\omega_h \delta \omega_4 c\theta_h \\ \nonumber + lk_f\omega_h^{2}s\theta_h \delta \theta_4 + 2k_m\omega_h \delta \omega_2 s\theta_h + k_m\omega_h^{2} c\theta_h \delta \theta_2 \\ \nonumber + 2k_m\omega_h \delta \omega_4 s\theta_h + k_m\omega_h^{2} c\theta_h \delta \theta_4
\end{eqnarray}
The tilt-rotor quadcopter achieves hover condition at zero tilt angle and acts like a conventional quadcopter. Thus, $\theta_h \xrightarrow{} 0$.
\begin{eqnarray}
\nonumber I_{xx} \medspace \delta\dot{p} = 2lk_f \omega_h (\delta \omega_2 - \delta \omega_4) + k_m\omega_h^{2} (\delta \theta_2 + \delta \theta_4) \\
\delta\dot{p} = \frac{2lk_f \omega_h (\delta \omega_2 - \delta \omega_4)}{I_{xx}} + \frac{k_m\omega_h^{2} (\delta \theta_2 + \delta \theta_4)}{I_{xx}} \label{eq16}
\end{eqnarray}
This expression yields the change in roll rate when the rotor speeds and rotor tilts are exercised simultaneously. It should be noted that the expression in \eqref{eq16} relates to the control of first vector element of the quaternion attitude. The change in angular speed of propellers ($\delta \omega_2 - \delta \omega_4 = \Delta \omega_{\phi}$) and rotor tilts ($\delta \theta_2 + \delta \theta_4 = \Delta \theta_{\phi}$)  provides the necessary control action for attitude change and it can be represented in state space form as shown in \eqref{eq18}.
\begin{eqnarray}
    \begin{bmatrix}
    \delta\dot{p} \\ \delta\dot{\mathbb{q_1}}
    \end{bmatrix} &=&  \begin{bmatrix}
   0 && 0 \\
   1 && 0
    \end{bmatrix} \begin{bmatrix}
   \delta{p} \\ \delta \mathbb{q_1}
    \end{bmatrix} + \begin{bmatrix}
\frac{lk_f \omega_h}{I_{xx}} && \frac{k_m\omega_h^{2}}{2I_{xx}} \\
  0 && 0
    \end{bmatrix} \begin{bmatrix}
    \Delta \omega_{\phi} \\ \Delta \theta_{\phi}
    \end{bmatrix} \label{eq18}
\end{eqnarray}
Similar expressions can be obtained about $y_b$ and $z_b$ axes. They can be easily accommodated in existing state space as shown from \eqref{eq19} to \eqref{eq22}
\begin{eqnarray}
    \delta\dot{q} &=& \frac{lk_f \omega_h (\delta \omega_3 - \delta \omega_1)}{I_{yy}} + \frac{k_m\omega_h^{2} (\delta \theta_3 + \delta \theta_1)}{2I_{yy}} \label{eq19}\\
    \begin{bmatrix}
    \delta\dot{q} \\ \delta\dot{\mathbb{q_2}}
    \end{bmatrix} &=&  \begin{bmatrix}
   0 && 0 \\
   1 && 0
    \end{bmatrix} \begin{bmatrix}
   \delta{q} \\ \delta \mathbb{q_2}
    \end{bmatrix} + \begin{bmatrix}
\frac{lk_f \omega_h}{I_{yy}} && \frac{k_m\omega_h^{2}}{2I_{yy}} \\
  0 && 0
    \end{bmatrix} \begin{bmatrix}
    \Delta \omega_{\theta} \\ \Delta \theta_{\theta}
    \end{bmatrix} \label{eq20}\\
 \delta\dot{r} &=& \frac{k_m\omega_h  \Delta \omega_{\psi}}{I_{zz}} + \frac{lk_f \omega_h^{2} \Delta \theta_{\psi}}{2I_{zz}} \label{eq21}\\
    \begin{bmatrix}
    \delta\dot{r} \\ \delta\dot{\mathbb{q_3}}
    \end{bmatrix} &=&  \begin{bmatrix}
   0 && 0 \\
   1 && 0
    \end{bmatrix} \begin{bmatrix}
   \delta{r} \\ \delta \mathbb{q_3}
    \end{bmatrix} + \begin{bmatrix}
\frac{k_m \omega_h}{I_{zz}} && \frac{lk_f\omega_h^{2}}{2I_{zz}} \\
  0 && 0
    \end{bmatrix} \begin{bmatrix}
    \Delta \omega_{\psi} \\ \Delta \theta_{\psi}
    \end{bmatrix} \label{eq22}    
\end{eqnarray}
where change in rotors' angular speed for yaw control $\Delta \omega_{\psi} = - \delta \omega_1 + \delta \omega_2 - \delta \omega_3 + \delta \omega_4$ and change in rotors' tilt angle for yaw control $\Delta \theta_{\psi} = - \delta \theta_1 - \delta \theta_2 + \delta \theta_3 + \delta \theta_4$. The quaternion operations in the controller are performed in normalized form. The desired quaternion $\hat{\mathbb{q}}^{des}$  and conjugate of the normalized quaternion $\hat{\mathbb{q}}$ are used to compute the error quaternion $\hat{\mathbb{q}}^{err}$ as shown in \eqref{eq23}.
\begin{equation}
\hat{\mathbb{q}}^{err} = \hat{\mathbb{q}}^{des} \bigotimes \hat{\mathbb{q}}^{*} \label{eq23}
\end{equation}
The error quaternion changes based on the required rotation for achieving the desired attitude. As, the UAV attains the desired attitude, the error quaternion becomes unit quaternion. It should be noted that the desired error quaternion $\hat{\mathbb{q}}_{err}^{des}$ is always a unit quaternion. Thus, the attitude control using quaternion feedback actually becomes a regulation problem in terms of quaternion error where the controller objective is to make the elements of the vector part of the error quaternion equal to zero. The change in propeller speeds and rotor tilt for achieving a desired orientation are shown in \eqref{eq24} and \eqref{eq25}, which are similar to expressions presented in \cite{fresk2013full}. Fig. \ref{fig2} shows the schematic of the attitude controller, here the outer loop generates commands for the inner loop. 
\begin{eqnarray}
\begin{bmatrix}
\Delta \omega_{\phi} \\ \Delta \omega_{\theta} \\ \Delta \omega_{\psi}
\end{bmatrix} &=& - k_{\mathbb{q}} \begin{bmatrix}
\Hat{\mathbb{q_1}}^{err} \\ \Hat{\mathbb{q_2}}^{err} \\ \Hat{\mathbb{q_3}}^{err}
\end{bmatrix} - k_{\omega} \begin{bmatrix}
p \\ q \\ r 
\end{bmatrix} \label{eq24}\\
\begin{bmatrix}
\Delta \theta_{\phi} \\ \Delta \theta_{\theta} \\\Delta \theta_{\psi}
\end{bmatrix} &=& - k_{\mathbb{q}}^{'}  \begin{bmatrix}
\Hat{\mathbb{q_1}}^{err} \\ \Hat{\mathbb{q_2}}^{err} \\ \Hat{\mathbb{q_3}}^{err}
\end{bmatrix} - k_{\omega}^{'}  \begin{bmatrix}
p \\ q \\ r 
\end{bmatrix} \label{eq25}
\end{eqnarray} 
Here, $k_{\mathbb{q}}, k_{\omega}, k_{\mathbb{q}}^{'}, k_{\omega}^{'}; \forall \in \mathds{R}^{3x3}$ are diagonal positive definite gain matrices.
\subsection{Lyapunov stability analysis}
We use a similar Lyapunov function discussed in \cite{junkins2009analytical} to analyze the stability of the tilt-rotor UAV attitude controller. 
\begin{lemma}
For the error dynamics given by
\begin{eqnarray}
\dot{\hat{\mathbb{q}}}_{err}&=& \frac{1}{2} \mathbb{\hat{q}}_{err} \bigotimes \begin{bmatrix}
0 \\ \Omega_{err}
\end{bmatrix}\label{errdyn_1} \\
\dot{\Omega}_{err} &=& \dot{\Omega} - \dot{\Omega}^{des} + \Omega \times \Omega^{des}\label{errdyn_2}
\end{eqnarray}
where $\Omega_{err} = \Omega - \Omega^{des} $, $\Omega^{des}$ is the desired angular velocity vector defined in the desired frame of reference and the vector elements for error quaternion $\mathbb{\hat{q}}_{err}$ are described by $\epsilon = [\Hat{\mathbb{q_1}}^{err}, ~\Hat{\mathbb{q_2}}^{err}, ~\Hat{\mathbb{q_3}}^{err}]^T $. The control law in \eqref{control_law} is globally stabilizing and reduces to expression similar to \eqref{eq24}, \eqref{eq25} for stabilization to a desired attitude, i.e. $\Omega^{des} \xrightarrow{}0$ as $\epsilon \xrightarrow{} 0 $. 
\begin{eqnarray}
    \tau \left(\Delta\omega_i, \Delta\theta_i\right ) &=& - k_{\mathbb{Q}} \epsilon - k_{\Omega} \Omega_{err} + I\dot{\Omega}^{des} \label{control_law} \\ \nonumber &-& I \left( \Omega \times \Omega^{des} \right) + \Omega \times I\Omega \qquad \forall i \in \lbrace \phi, \theta, \psi \rbrace 
\end{eqnarray}
\end{lemma}
\textit{Proof:} Consider the Lyapunov function candidate
\begin{equation}
\nonumber    V = k_{\mathbb{Q}} (\hat{\mathbb{q}}^{err} - \hat{\mathbb{q}}_{err}^{des})^{T} ( \hat{\mathbb{q}}^{err} - \hat{\mathbb{q}}_{err}^{des})
+ \frac{1}{2} \Omega_{err}^T I \Omega_{err}
\end{equation}
The time derivative of the Lyapunov candidate is computed and substituting for $\dot{\hat{\mathbb{q}}}_{err}$, $\dot{\Omega}_{err}$ and $\dot{\Omega}$ from \eqref{errdyn_1}, \eqref{errdyn_2} and \eqref{eq1}.
\begin{eqnarray}
\nonumber \dot{V} &=& 2k_{\mathbb{Q}} (\hat{\mathbb{q}}^{err} - \hat{\mathbb{q}}_{err}^{des})^{T} (\dot{\hat{\mathbb{q}}}^{err} - \dot{\hat{\mathbb{q}}}_{err}^{des}) + \Omega_{err} I\dot{\Omega}_{err} \\ 
\nonumber \dot{V} &=& k_{\mathbb{Q}} \Omega_{err}^T [f(\hat{\mathbb{q}}^{err})]^{T} (\hat{\mathbb{q}}^{err} - \hat{\mathbb{q}}_{err}^{des}) \\ \nonumber &+& \Omega_{err}^T \left( I \dot{\Omega} - I \dot{\Omega}^{des} + I \left( \Omega \times \Omega^{des} \right)  \right)
\end{eqnarray}
where the definition of $[f(\hat{\mathbb{q}}^{err})]$ is identical to that explained in \cite{junkins2009analytical}, $[f(\hat{\mathbb{q}}^{err})]^{T} \hat{\mathbb{q}}^{err} = 0 $ and $[f(\hat{\mathbb{q}}^{err})]^{T} \hat{\mathbb{q}}_{err}^{des} = \epsilon$. The vector part of error quaternion is given by $\epsilon = [\Hat{\mathbb{q_1}}^{err}, ~\Hat{\mathbb{q_2}}^{err}, ~\Hat{\mathbb{q_3}}^{err}]^T $.
\begin{eqnarray}
\nonumber \dot{V} &=& k_{\mathbb{Q}} \Omega_{err}^T \epsilon + \Omega_{err}^T \left( -\Omega \times I\Omega + \tau - I \dot{\Omega}^{des} + I ( \Omega \times \Omega^{des}) \right )
\end{eqnarray}
Substituting \eqref{control_law} in the Lyapunov time derivative equation, we obtain the following result after simplification
\begin{equation}
\nonumber \dot{V} = - k_\Omega \Omega_{err}^{T} \Omega_{err} \leq 0
\end{equation}
where $k_{\mathbb{Q}}, k_{\Omega}; \forall \in \mathds{R}^{3x3}$ are diagonal positive definite gain matrices. The control torque $\tau$ is a function of variation in propeller angular speeds and tilt angles as shown in \eqref{eq18}, \eqref{eq20}, \eqref{eq22}. The Lyapunov function here considers $\Omega_{err}$, but the final implementation reduces to the expressions shown in \eqref{eq24}, \eqref{eq25} upon simplification. Here, $\Omega^{des} = - k_{\mathbb{Q}} \epsilon$, and $\Omega^{des} \xrightarrow{}0$ as $\epsilon \xrightarrow{} 0 $. The Lyapunov candidate is unbounded, the stability properties hold globally. It is straightforward to show asymptotic stability using invariance principles.

\subsection{Position Control}
Here, the position controller of the UAV is developed which generates quaternion commands for the inner attitude controller. The tilting motion of rotors causes acceleration along $xy$-axes. Hence the longitudinal and lateral motion of the system can also be controlled using rotor tilts.
The position errors ($e_x, e_y, e_z$), velocity errors ($\dot{e_x}, \dot{e_y}, \dot{e_z}$) and error integrals are utilized by the outer PID controller loop as shown in (\ref{eq271}) to compute rotor-tilt commands $\Delta\theta_i; \forall {i}\in \lbrace{x,y}\rbrace $ and desired accelerations commands $\ddot{r_i}^{des}; \forall {i}\in \lbrace{x,y,z}\rbrace$.
\begin{eqnarray}
\nonumber \ddot{r_x}^{des} &=& k_{p_x}e_x + k_{i_x}\int{e_x} dt +  k_{d_x}\dot{e_x} \\ [-2pt]
\ddot{r_y}^{des} &=& k_{p_y}e_y + k_{i_y}\int{e_y} dt +  k_{d_y}\dot{e_y} \label{eq271} \\ [-2pt]
\nonumber \ddot{r_z}^{des} &=& k_{p_z}e_z + k_{i_z}\int{e_z} dt +  k_{d_z}\dot{e_z}  + g \\ [-2pt]
\nonumber \Delta\theta_{x} &=& k_{p_{\theta_x}}e_x + k_{i_{\theta_x}} \int{e_x} dt +  k_{d_{\theta_x}}\dot{e_x} \\[-2pt]
\nonumber \Delta\theta_{y} &=& k_{p_{\theta_y}}e_y + k_{i_{\theta_y}}\int{e_y} dt +  k_{d_{\theta_y}}\dot{e_y}
\end{eqnarray} 
Here, $k_{p_i}$, $k_{i_i}$, and $k_{d_i}$ $\forall {i}\in \lbrace{x,y,z}\rbrace$, $k_{p_{\theta_i}}$, $k_{i_{\theta_i}}$, and $k_{d_{\theta_i}}$ $\forall {i}\in \lbrace{x,y}\rbrace$ are the proportional, integral and derivative gains for the position controller. The angular speed required for individual propeller motors necessary for hovering and motion along the $z-$axis is given by \eqref{eq81}.
\begin{eqnarray} 
\omega_h=\sqrt{\dfrac{m\ddot{r_z}^{des}}{k_f(c\theta_1 + c\theta_2 + c\theta_3 + c\theta_4 )}}   \label{eq81}
\end{eqnarray}
The body accelerations are represented by $a_{b_i}$ $\forall {i}\in \lbrace{x,y,z}\rbrace$ for the system. The objective is to determine the quaternion which will align the body acceleration vector $a_{b} = [a_{b_x} \quad a_{b_y} \quad a_{b_z}]^{T}$ along the desired inertial acceleration vector $\ddot{r}^{des} = [\ddot{r_x}^{des} \quad \ddot{r_y}^{des} \quad \ddot{r_z}^{des}]^{T}$ from \eqref{eq271}. This is achieved by normalizing the two vectors as $\hat{a_{b}} = norm[a_b]$ , $\hat{a_i} = norm[\ddot{r}^{des}]$ and computing the required rotation $\Theta$ and axis of rotation $\hat{\textbf{n}}$. The cosine, sine of the rotation angles and the axis of rotation can be calculated with vector multiplication operations \cite{kehlenbeck2014quaternion}. The resulting quaternion is denoted by $\Tilde{\mathbb{q}}$.
\begin{eqnarray}
\Tilde{\mathbb{q}} = \frac{1}{\sqrt{2(1 + \hat{a_b}^{T} \hat{a_i})}} \begin{bmatrix}
    1 + \hat{a_b}^{T} \hat{a_i}\\\\
    \hat{a_b} \times \hat{a_i}
    \end{bmatrix} \label{eq291}
\end{eqnarray}
The commanded orientation $\Tilde{\mathbb{q}}$ can be corrected for desired yaw angle as discussed in \cite{cutler2015analysis} which yields $\mathbb{q}^{des}$. The desired quaternion is further normalized to $\hat{\mathbb{q}}^{des}$ for commanding the attitude controller loop. The control allocation matrix for the entire system is shown in \eqref{eq311}. The attitude control elements are derived from the state space formulation discussed earliar. The rotor speeds are represented by $\omega_i = \omega_h + \Delta \omega_j, \medspace \forall i \in \lbrace 1,2,3,4 \rbrace, j \in \lbrace \phi, \theta, \psi \rbrace $. Similarly, the rotor tilt angles are given as $\theta_i = \theta_h + \Delta \theta_j, \medspace \forall i \in \lbrace 1,2,3,4 \rbrace, j \in \lbrace \phi, \theta, \psi, x, y \rbrace $.
\begin{eqnarray}
\begin{bmatrix}
\Delta \omega_1\\
\Delta \omega_2\\
\Delta \omega_3\\
\Delta \omega_4\\
\Delta \theta_1\\
\Delta \theta_2\\
\Delta \theta_3\\
\Delta \theta_4
\end{bmatrix} = \begin{bmatrix}
0 & -1 & -1 & 0 & 0 & 0 & 0 & 0  \\
1 & 0 & 1 & 0 & 0 & 0 & 0 & 0   \\
0 & 1 & -1 & 0 & 0 & 0  & 0 & 0  \\
-1 & 0 & 1 & 0 & 0 & 0 & 0 & 0 \\
0 & 0 & 0 & 0 & 1 & -1 & 0 & -1   \\
0 & 0 & 0 & 1 & 0 & -1 & 1 & 0  \\
0 & 0 & 0 & 0 & 1 & 1 & 0 & -1  \\
0 & 0 & 0 & 1 & 0 & 1 & 1 & 0 \\
\end{bmatrix}
\begin{bmatrix}
\Delta\omega_\phi\\
\Delta\omega_\theta\\
\Delta\omega_\psi\\
\Delta\theta_{\phi}\\
\Delta\theta_{\theta}\\
\Delta\theta_{\psi}\\
\Delta\theta_{x}\\
\Delta\theta_{y}
\end{bmatrix} \label{eq311}
\end{eqnarray}

\section{Numerical Simulations and Results}
\begin{figure}[]
	\centering
	\includegraphics[scale=0.45]{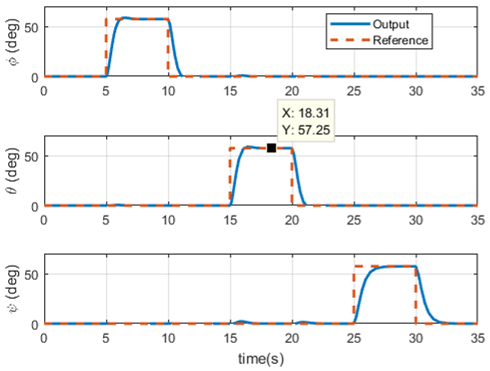}
    \caption{Variation in Euler angles} \vspace{-5mm}
    \label{fig3}
\end{figure}

\begin{figure}[]
	\centering
	\includegraphics[scale=0.45]{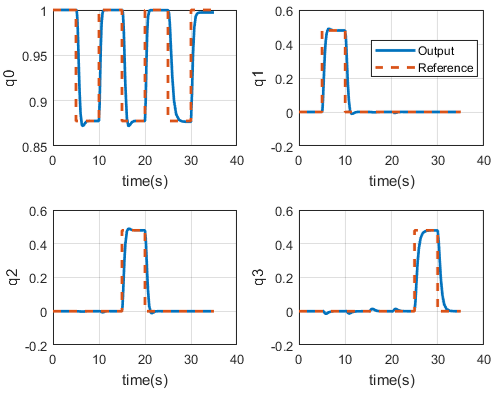}
    \caption{Variation in quaternion elements} \vspace{-8mm}
    \label{fig4}
\end{figure}

\begin{figure}[]
	\centering
	\includegraphics[scale=0.45]{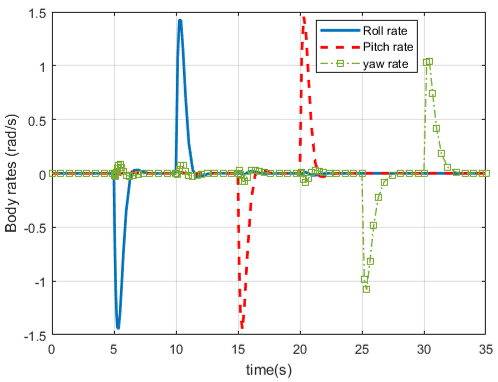}
    \caption{Body rates about $x_by_bz_b$-axes} \vspace{-5mm}
    \label{fig6}
\end{figure}

Here, the proposed controller is validated by numerical simulations. The mathematical model of the UAV and controller are developed in MATLAB and Simulink R2017a. The parameters used in the simulations are $m=1.56kg,~ l=0.12m,~ k_f=2.2e-4 Ns/rad,~ k_m=5.4e-6Ns/rad,~ I_{xx} = I_{yy}= 0.0449kgm^2,~ I_{zz} = 0.0899kgm^2$. The first simulation considers application of step input to the roll, pitch and yaw axes of the UAV. The roll command of $1rad$ is issued at $t = 5s$. Similarly, the pitch command is issued at $t = 15s$ and yaw command is issued at $t = 25 s$. Figure \ref{fig3} and \ref{fig6} show the variation of Euler angles and body rates w. r. t. step inputs. It can be seen in Fig. \ref{fig4} that the second element of quaternion changes when roll step input is commanded to the UAV. Similarly, third and fourth element of quaternion respond to pitch and yaw step inputs respectively. Further, we validate the flight controller by simulating a way point navigation mission. The UAV is initialized at the origin and commanded to visit a predefined set of way points at a height of $5m$. The set of way points are $[5, \thickspace 5]$, $[5, \thickspace 10]$, $[10,\thickspace 10]$, $[15,\thickspace20]$, $[20,\thickspace 20]$.  The position controller generates necessary rotor-tilt and desired quaternion commands to minimize the position error. The three dimensional trajectory followed by the UAV is shown in figure \ref{fig10}. The UAV visits all way-points by maintaining the desired height. The tilt-rotor UAV has redundancy in control of position and orientation because the position control is achieved by changing the UAV orientation as well as rotor-tilt angles. Figure \ref{fig11} shows the variation of angular speeds of the UAV propellers. Similarly, figure \ref{fig13} shows the variation of rotor-tilt angles. The error-quaternion elements are shown in figure \ref{fig12}. The error-quaternion elements change while the UAV is navigating between way-points and they converge to the unit quaternion as the UAV reaches goal position.
\begin{figure}[]
	\centering
	\includegraphics[scale=0.45]{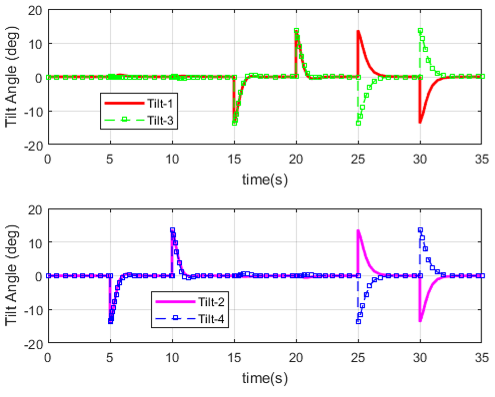}
    \caption{Variation in rotor tilt angles} \vspace{-5mm}
    \label{fig7}
\end{figure}
\begin{figure}[]
	\centering
	\includegraphics[scale=0.45]{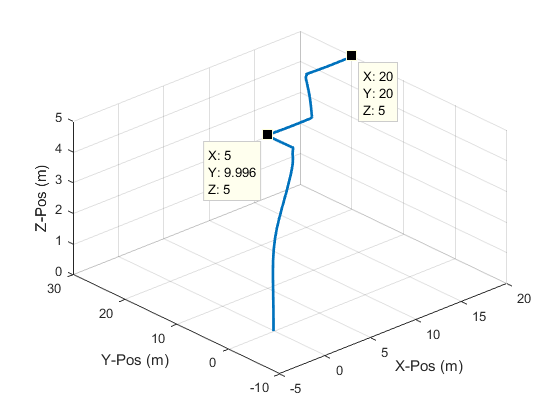}
    \caption{Trajectory for WPN} \vspace{-5mm}
    \label{fig10}
\end{figure}
\begin{figure}[]
	\centering
	\includegraphics[scale=0.45]{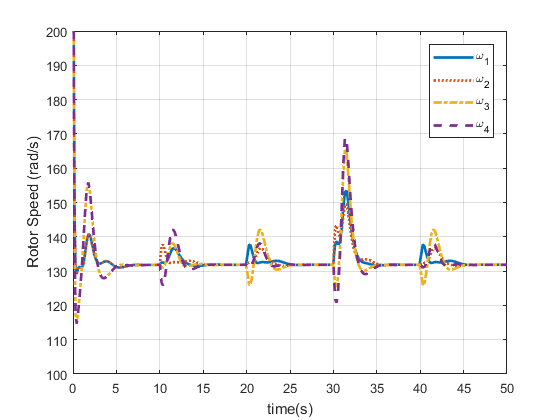}
    \caption{Variation of rotor speeds during WPN} \vspace{-5mm}
    \label{fig11}
\end{figure}
\begin{figure}[]
	\centering
	\includegraphics[scale=0.45]{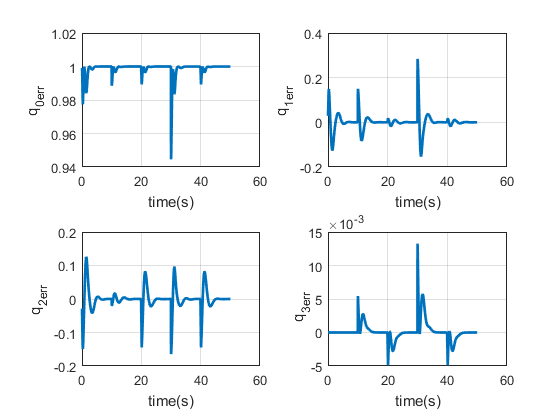}
    \caption{Quaternion error elements during WPN} \vspace{-5mm}
    \label{fig12}
\end{figure}
\begin{figure}[]
	\centering
	\includegraphics[scale=0.45]{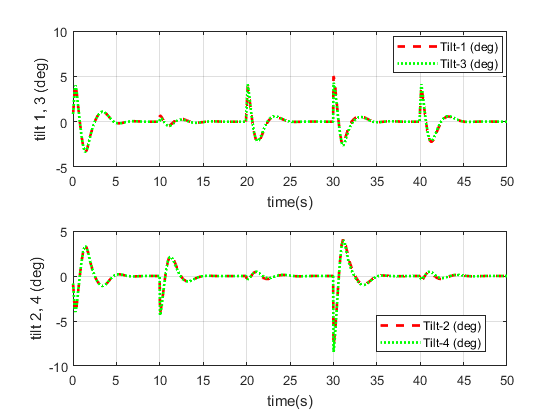}
    \caption{Variation in rotor-tilt angles during WPN} \vspace{-5mm}
    \label{fig13}
\end{figure}

\section{Conclusion}
In this paper, position and attitude controller for the tilt-rotor quadcopter with quaternion feedback was presented. The UAV dynamics for translational and rotational motion were shown. Taylor series expansion about the hover condition was used to derive the necessary control allocation for the system. Lyapunov stability analysis of the attitude controller was presented. The performance of the quaternion feedback attitude controller was shown for reference attitude tracking. The inner quaternion feedback loop was commanded using an external position controller for a way-point mission. The complete control allocation and simulations were presented for achieving way-point navigation. Future work will involve experimental validation of the proposed flight controller and more studies will be conducted to exploit the redundant control inputs for achieving fault-tolerant control during flight. 
\bibliographystyle{IEEEtran}
\bibliography{refdatabase}
\end{document}